\documentclass[conference, 10pt]{IEEEtran}
\IEEEoverridecommandlockouts

\usepackage{color}
\usepackage{theorem}
\usepackage{times,amsmath,epsfig}
\usepackage{amssymb}
\usepackage{subcaption}
\usepackage{cite}
\usepackage{url}
\usepackage[capitalise]{cleveref}
\usepackage[ruled,linesnumbered]{algorithm2e}

\input{mysymbol.sty}

\usepackage{tikz, pgfplots}
\usetikzlibrary{shapes,arrows}
\pgfplotsset{compat=1.17}
\pgfplotstableset{col sep=semicolon}
\usepackage{moresize} 
\usepgfplotslibrary{groupplots}

\tikzset{every mark/.append style={scale=1.6, solid}, font=\small}
\pgfplotsset{
    width=1\textwidth,
    height=5.5cm,
    legend style={
        font=\ssmall ,  
        inner xsep=1pt,
        inner ysep=1pt,
        nodes={inner sep=1pt}},
    legend cell align=left,
    every axis/.append style={line width=.5pt},
 	every axis plot/.append style={line width=2.25pt},
 	every axis y label/.append style={yshift=-4pt}
}

\def\bchkS{{\ensuremath{\mathbf{\check S}}}}

\title{Robust Graph Neural Network based on Graph Denoising}
\author{Victor M. Tenorio,
        Samuel Rey,
        and~Antonio G. Marques \\
        Department of Signal Theory and Communications\\
        King Juan Carlos University, Madrid, Spain\\
        Emails: \{victor.tenorio,samuel.rey.escudero,antonio.garcia.marques\}@urjc.es.
        
\thanks{This work was supported in part by
the Spanish AEI under Grants PID2019-105032GBI00, PID2022-136887NBI00 and FPU20/05554, in part by the Young Researchers
R\&D Project under ref. num. F861 (CAM and URJC) and in part by the Autonomous Community of Madrid within the ELLIS Unit Madrid framework.}}

\begin{document}
%
\maketitle
\begin{abstract}
Graph Neural Networks (GNNs) have emerged as a notorious alternative to address learning problems dealing with non-Euclidean datasets.
However, although most works assume that the graph is perfectly known, the observed topology is prone to errors stemming from observational noise, graph-learning limitations, or adversarial attacks.
If ignored, these perturbations may drastically hinder the performance of GNNs.
To address this limitation, this work proposes a robust implementation of GNNs that explicitly accounts for the presence of perturbations in the observed topology. 
For any task involving GNNs, our core idea is to i) solve an optimization problem not only over the learnable parameters of the GNN but also over the true graph, and ii) augment the fitting cost with a term accounting for discrepancies on the graph. 
Specifically, we consider a convolutional GNN based on graph filters and follow an alternating optimization approach to handle the (non-differentiable and constrained) optimization problem by combining gradient descent and projected proximal updates.
The resulting algorithm is not limited to a particular type of graph and is amenable to incorporating prior information about the perturbations.
Finally, we assess the performance of the proposed method through several numerical experiments.
\end{abstract}

\section{Introduction}\label{S:Introduction}
For a few years now, graph neural networks (GNNs) have emerged as a prominent alternative to handle contemporary data in a wide variety of fields~\cite{scarselli2009gnn,bronstein2017geometricdeeplearning,wu2020comprehensive,bronstein2021geometric,rey2021overparametrized}.
The rising popularity of these architectures can be largely attributed to two pivotal factors.
First, we are experiencing a data deluge where vast amounts of data are generated and stored, hence propelling the development of data-driven alternatives. Second, contemporary data is not only becoming more abundant but also more heterogeneous and intricate.
In light of these dynamics, GNNs draw inspiration from the approaches put forth in fields like graph signal processing (GSP) and graphical models, where the graph topology is leveraged to deal with the underlying irregular domain inherent in contemporary data~\cite{kolaczyk2009book,shuman2013emerging,djuric2018cooperative,marques2020editorial}.
Indeed, by accounting for the information encoded in the graph topology, GNNs provide state-of-the-art results in a gamut of applications such as drug discovery, recommender systems, signal denoising, or traffic state predictions~\cite{kearnes2016molecular,ying2018graph,guo2019attention,zhou2020graph,rey2022untrained}, to name a few.

Due to the outstanding performance of GNNs, numerous alternatives for integrating the graph structure into their architecture have emerged, typically operating under the foundational assumption that the topology of the graph is precisely known.
Nevertheless, this assumption is unlikely to hold in many practical setups, where graphs may suffer from \emph{perturbations}.
These perturbations can arise in various contexts.
When networks are explicitly provided, perturbations may be due to observational noise and errors.
On the other hand, when graphs represent (statistical) pairwise relationships among observed variables, they must be inferred from the available data, introducing another layer of potential perturbations.
On top of this, the observed graph may be subject to adversarial attacks~\cite{dai2018adversarial}.
Intuitively, since GNNs exploit the information encoded in the graph topology to process the data, ignoring the presence of perturbations in the graph can drastically hinder their performance.

Accounting for the influence of graph perturbations is a challenging problem that has been attracting increasing attention.
Initially,~\cite{ceci2020graph} studied how perturbations affected the spectrum of the graph Laplacian.
Then,~\cite{kenlay2021stability,nguyen2022stability} first characterized the stability of linear graph filters when the graph topology presents errors and, later on,~\cite{levie2021transferability,ruiz2021graph,keriven2020convergence} characterized the stability and transferability of GNNs.
Although these works showed that the discrepancies in the output of a GNN due to perturbations are upper-bounded, this bound grows i) exponentially with the number of layers and ii) linearly with the discrepancies in the graph filters due to perturbations, which are shown to grow exponentially with the order of the filter~\cite{rey2023robust}.
Therefore, a different line of research is concerned with solving graph-related tasks from a robust perspective.
In the context of GSP,~\cite{ceci2020_semtls} approached the robust inference of graph signals by combining total least squares and structural equation models, and~\cite{natali2020topology,rey2021robust,rey2023robust} tackle the robust identification of graph filters, first assuming that perturbations do not affect the support of the graph and then under a more general perturbation model.
Moving on to GNNs,~\cite{tenorio2021robust} introduces a novel graph-aware operator that is more resilient to perturbations than classical polynomials of the adjacency matrix.
Differently,~\cite{jin2020graph} addresses the task of node classification when the observed topology is perturbed by inferring the topology of the true unknown graph while training the GNN.
Although relevant, the proposed approach assumes that the target graph has a low-rank adjacency matrix and that node features are smooth on the graph, which may be too restrictive for general settings.
Moreover, the method does not account for any prior information about the particular type of perturbation, and the selected convolutional layer may lead to oversmoothing if a deep architecture is considered, limiting the number of layers.

In contrast to previous methods, we assume that the observed topology of the graph is a perturbed version of the true (unknown) topology, and develop a general method to train GNNs that alleviates the impact of errors in the topology.
To that end, we consider the graph structure as an optimization variable and leverage the assumption that it should be \emph{close} to the perturbed observation.
However, rather than being constrained to a particular type of graph, the proposed model can accommodate any (statistical) prior knowledge about the observed perturbations and the topology of the graph.
On top of this, we consider a generalization of the convolutional layer from~\cite{kipf2016semi} where the vertex-based convolution is modeled by a bank of learnable graph filters, hence avoiding the oversmoothing issue and decoupling the depth and the radius (neighborhood) of the architecture.
Then, we jointly estimate the learnable parameters of the GNN and the true (unknown) graph topology.
Because the estimation of the graph topology is likely to include non-differentiable terms in the objective function, we approach the optimization problem with an alternating minimization algorithm combining gradient descent steps for the differentiable terms and projected proximal updates for the non-differentiable elements.
As a result, the proposed method can efficiently harness prior information about the graph and the perturbations.

\section{Notation and Fundamentals}
This section covers basic notions about GSP, introduces the particular GNN that will be considered through this work, and briefly establishes the notation for the node classification task.

\vspace{2mm}
\noindent\textbf{Fundamentals of GSP.}
Let $\ccalG = (\ccalV, \ccalE)$ denote a graph composed of the set of $N$ nodes $\ccalV$ and the set of edges $\ccalE$.
The topology of $\ccalG$ is encoded in the adjacency matrix $\bbA \in \reals^{N \times N}$, which is a sparse matrix such that $A_{ij}=0$ if and only if $(i,j) \not\in \ccalE$.
The value $A_{ij}$ of the non-zero entries captures the strength of the link between the nodes $i$ and $j$. Graph signals are a particular type of signal defined on the nodes of the graph $\ccalV$.
A graph signal can be conveniently represented by a vector $\bbx \in \reals^N$ with the entry $x_i$ denoting the signal value at node $i$.
Of particular interest when processing graph signals is the graph-shift operator (GSO), a linear operator applied to graph signals that captures the topology of the graph~\cite{shuman2013emerging}.
The GSO is represented by the matrix $\bbS \in \reals^{N \times N}$ whose entries satisfy that $S_{ij} \neq 0$ only if $i = j$ or $(i,j) \in \ccalE$.
Typical choices for the GSO include the adjacency matrix $\bbA$, or the combinatorial graph Laplacian~\cite{shuman2013emerging}. 
Finally, a graph filter is a graph-aware linear mapping between graph signals, which can be conveniently represented as a polynomial of the GSO $\bbH = \sum_{r=0}^{R-1} h_r \bbS^r$, where $\bbh = [h_0, ..., h_{R-1}]$ is the vector of filter coefficients.

\vspace{2mm}
\noindent\textbf{GNNs based on graph filters.}
Generically, we represent a GNN as a parametric non-linear function $f_{\bbTheta}(\bbX|\bbS): \reals^{N \times F_i} \to \reals^{N \times F_o}$ that depends on the graph structure encoded in $\bbS$. The parameters of the architecture are collected in $\bbTheta$, and $\bbX \in \reals^{N \times F_i}$ represents the input of the network with $F_i$ features.
Among the different possibilities to define a GNN, we focus on an architecture whose layers combine learnable linear mappings in the form of a bank of graph filters and point-wise non-linearities~\cite{ruiz2021graph}.
The output of such architecture is given by the following recursion
\begin{equation}\label{eq:GNN}
     \bbX_\ell = \sigma_\ell \left( \sum_{r=0}^{R-1} \bbS^r \bbX_{\ell - 1} \bbTheta_{\ell,r} \right),
\end{equation}
where $\bbTheta=\{\{\bbTheta_{\ell,r}\}_{r=0}^{R-1}\}_{\ell=1}^L$ represent the learnable parameters of the network, i.e. the coefficients of the graph filters in the bank; $\sigma_\ell$ is a non-linear function applied pointwise and $\bbX_\ell$ are the intermediate graph signals learned at layer $\ell$.
Note that the linear transformation based on graph filters in \eqref{eq:GNN} endows the GNN with important benefits.
First, it decouples the depth of the architecture and the range (neighborhood) of the GNN.
Second, since the coefficients of the graph filters are learnable parameters the architecture is not restricted to low-pass filters, hence avoiding the oversmoothing issue~\cite{chen2020measuring}.

\vspace{2mm}
\noindent\textbf{Node classification.}
Although the approach proposed in this work can be applied to any problem involving GNNs, for the sake of simplicity let us focus on the task of semi-supervised node classification.
Denote by $\bbX := [\bbx_1,...,\bbx_F]\in \reals ^{N \times F}$ the matrix collecting all the node features, and let the set $\ccalY := \{y_1,...,y_N\}$ collect the set of all node labels, from which only the subset of $M$ labels $\ccalY_{train} := \{y_1,...,y_M\}$ is known.
In the classical setting when the graph is perfectly known, the node classification task is solved by fitting the weights of the GNN to solve the optimization problem
\begin{equation}\label{eq:classical_GNN}
    \min_{\bbTheta} \ccalL (f_{\bbTheta} ( \bbX |\bbS), \ccalY_{train}) \;\;\;\text{with}\;\; \bbTheta=\{\{\bbTheta_{\ell,r}\}_{r=0}^{R-1}\}_{\ell=1}^L,
\end{equation}
where $f_{\bbTheta} (\bbX |\bbS)$ represents the considered GNN and $\ccalL$ represents an appropriate loss function (e.g., the cross-entropy loss).
However, in many real-world problems, the true $\bbS$ might not be available.
How to learn the parameters of the architecture when only a perturbed observation of the GSO is available is the subject of the following section.

\section{Robust GNNs based on Graph Denoising}
This section discusses how to design a GNN that is robust to imperfections in the topology of the graph.
To that end, we focus on the task of semi-supervised node classification and recall that the node features and node labels are denoted as $\bbX\in \reals ^{N \times F}$ and $\ccalY= \{y_1,...,y_N\}$, respectively.
Moreover, let $\barbS \in \reals^{N \times N}$ be a \emph{perturbed observation} of the true unknown GSO, and consider the additive perturbation model
\begin{equation}
    \barbS = \bbS + \bbDelta,
\end{equation}
where $\bbDelta$ represents a perturbation matrix whose structure depends on the particular type of perturbation.
Relevant examples of perturbations include creating and destroying edges, or noisy weights.
In the first case, assuming an unweighted graph, $\Delta_{ij}=1$ represents that the perturbation is creating an edge between nodes $i$  and $j$ while $\Delta_{ij}=-1$ indicates that the edge $(i,j)$ is being destroyed. 
On the other hand, when the perturbations represent uncertainty over the edge weights, the support of $\bbDelta$ will match the support of $\bbS$ and the non-zero entries of $\bbDelta$ will be sampled from a distribution modeling the observation noise~\cite{rey2023robust}.
Finally, because $\barbS$ represents a perturbed version of $\bbS$, the distance between the true and the observed GSO is assumed to be small according to some metric $d(\bbS, \barbS)$.
This assumption captures that $\barbS$ contains some information about the true GSO, thus ensuring the tractability of the problem.

With the previous definitions in place, our goal is to address the node classification task from a robust perspective by jointly estimating the missing node labels and recovering the original $\bbS$.
To that end, we consider the following problem
\begin{alignat}{2}\label{eq:model1}
    \!&  \!&&\ \min_{\bbTheta,\bbS} \ccalL\left( f_{\bbTheta} ( \bbX | \bbS ), \ccalY_{train} \right) + \alpha d(\bbS,\barbS) + \lambda \gamma(\bbS) \nonumber\\ 
    \!&  \!&&\ \mathrm{\;\;s. \;to:  \bbS \in \ccalS}.
\end{alignat}
The function $\gamma(\bbS)$ allows us to incorporate prior knowledge about the true GSO (e.g., $\bbS$ being sparse or low rank). Similarly, the particular choice of $d(\bbS,\barbS)$ is determined by the model assumed for the perturbation $\bbDelta$, so it can capture information available about the perturbations.
Then, the set of convex constraints $\bbS \in \ccalS$ ensures that the learned GSO belongs to a desired family (e.g., the set of adjacency matrices with non-zero diagonal entries).

While the method put forth in \eqref{eq:model1} may accommodate any functions $d(\bbS,\barbS)$ and $\gamma(\bbS)$, 
for the sake of simplicity, in the remainder of the paper we assume that $\barbS$ is the result of creating and/or destroying edges in $\bbS$.
Therefore, we set $d(\bbS,\barbS) = \| \bbS - \barbS \|_1$ (note that the $\ell_1$ is the standard convex surrogate of the $\ell_0$ pseudonorm, which is the workhorse choice to promote sparsity).
Moreover, we will only assume that $\bbS$ is sparse and set $\gamma(\bbS) = \| \bbS \|_1$.
Nevertheless, we note that both functions can be readily replaced by other alternatives.

Despite the aforementioned benefits, considering $\bbS$ as an optimization variable renders the optimization problem more challenging to solve.
Next, we discuss an algorithmic approach to alleviate this limitation.

\subsection{Algorithmic implementation of robust GNNs}
The prevailing approach to fit the parameters of a GNN involves minimizing a desired loss function using stochastic gradient descent (SGD).
However, the optimization problem in \eqref{eq:model1} is a challenging constrained optimization problem that i) involves non-differentiable terms and ii) the matrix of input features is multiplied both from the right and from the left by optimization variables [see \eqref{eq:GNN}].
As a result, standard SGD methods might incur in difficulties estimating both $\bbTheta$ and $\bbS$.
To circumvent this limitation, we follow an alternating optimization approach where each step involves solving a simpler optimization problem.
More precisely, the resulting algorithm solves the two following subproblems for $t=0,...,T_{max}-1$ iterations.

\noindent\textbf{Step 1.}
We estimate the block of variables collected in $\bbTheta$ while the current estimate of the GSO, $\bbS^{(t)}$, remains fixed.
This results in the optimization problem 
\begin{equation}\label{eq:step1}
    \bbTheta^{(t+1)} = \argmin_{\bbTheta} \ccalL( f_{\bbTheta} (\bbX | \bbS^{(t)} ), \ccalY_{train} ),
\end{equation}
which amounts to the classical minimization for training a GNN considering $\bbS^{(t)}$ as the real GSO [see \eqref{eq:classical_GNN}].
This step can be solved via SGD and backpropagation.

\noindent\textbf{Step 2.}
Now we estimate the GSO $\bbS$ while the weights of the architecture collected in $\bbTheta^{(t+1)}$ remain fixed.
This results in the optimization problem
\begin{alignat}{2}\label{eq:step2}
    \!&    \bbS^{(t+1)} = \argmin_{\bbS\in \ccalS} \!&&\   \ccalL\left( f_{\bbTheta^{(t+1)}} ( \bbX|\bbS ), \ccalY_{train} \right)  \nonumber \\
    \!&  \!&&\ + \alpha \| \bbS - \barbS \|_1 + \lambda\| \bbS \|_1. 
\end{alignat}
Different from the previous step, the estimation of $\bbS^{(t+1)}$ involves a constrained optimization problem with non-differentiable terms, which prevents us from directly applying SGD.
To circumvent this situation, we employ a projected proximal gradient descent algorithm.
Then, to solve \eqref{eq:step2} we consider a nested iterative process where, after initializing the variable $\bchkS^{(0)} = \bbS^{(t)}$, at each inner iteration $\tau$, we perform the following sequence of operations
\begin{align}
    &\dot{\bbS} = \bchkS^{(\tau)} - \eta\nabla_\bbS \ccalL \left( f_{\bbTheta^{(t+1)}}(\bbX 
    | \bchkS^{(\tau)}), \ccalY_{train} \right) \label{eq:ppm_first}, \\
    &\ddot{\bbS} =  \mathrm{prox}_{\eta\lambda\|\cdot\|_1}(\dot{\bbS}), \\
    &\dddot{\bbS} =  \mathrm{prox}_{\eta\alpha d(\cdot, \barbS)}(\ddot{\bbS}), \\
    &\bchkS^{(\tau + 1)} = \Pi_\ccalS(\dddot{\bbS}) \label{eq:ppm_last}.
\end{align}
Put in words, we perform a gradient step of the differentiable terms followed by a proximal update for the non-differentiable terms.
Here, $\mathrm{prox}_{\eta\lambda\|\cdot\|_1}(\cdot)$ and $\mathrm{prox}_{\eta\alpha d(\cdot, \barbS)}(\cdot)$ respectively correspond to the proximal update of $\lambda\| \bbS \|_1$ and $\alpha\| \bbS - \barbS \|_1$.
The proximal operator associated with $\| \bbS \|_1$ is the soft-thresholding operator given by
\begin{equation}
    \mathrm{prox}_{\lambda\|\cdot\|_1}(\bbS) = \sign(\bbS)\circ(|\bbS| - \lambda)^+,
\end{equation}
where $\circ$ is the Hadamard (entry-wise) product, the absolute value is applied in an entry-wise fashion, and $(\cdot)^+$ denotes the operator $(x)^+=\max(0,x)$.
Similarly, the proximal operator of $d(\bbS, \barbS) = \|\bbS - \barbS\|_1$ is the shifted version of the soft-thresholding, which is given by
\begin{equation}
    \mathrm{prox}_{\alpha \| \cdot - \bar{S}_{ij} \|_1}(S_{ij}) = 
    \left\{\hspace{-2mm} \begin{array}{cl}
        S_{ij} - \alpha 
        &  \mathrm{if} \; S_{ij} - \bar{S}_{ij} > \alpha \\
        S_{ij} + \alpha 
        & \mathrm{if} \; S_{ij} - \bar{S}_{ij} < -\alpha, \\
        \bar{S}_{ij} & \mathrm{otherwise}.
    \end{array} \right.
\end{equation}
Finally, $\Pi_\ccalS(\cdot)$ denotes the projection onto the convex set $\ccalS$.

The overall algorithm to train the robust GNN is summarized in Algorithm~\ref{alg:robust_gnn}.
It is worth recalling that, although we focused on the case when $\gamma(\bbS)=\|\bbS\|_1$ and $d(\bbS,\barbS)=\|\bbS-\barbS\|_1$ for the sake of simplicity, the proposed algorithm can be easily modified to account for other functions of interest by incorporating any additional differentiable term in \eqref{eq:ppm_first} and then solving the appropriate proximal operators of the non-differentiable terms.

\begin{algorithm}[tb]
\SetKwInput{Input}{Input}
\SetKwInOut{Output}{Output}
\Input{$\bbX$, $\ccalY_{train}$, $\barbS$}
\Output{$\hbTheta$, $\hbS$}
\SetAlgoLined
Initialize $\bbS^{(0)}$. \\
\For{$t=0$ \KwTo $T_{max}-1$}{
    Compute $\bbTheta^{(t+1)}$ by solving \eqref{eq:step1} fixing $\bbS^{(t)}$. \\
    Initialize $\bchkS^{(0)} = \bbS^{(t)}$ \\
    \For{$\tau = 0$ \KwTo $\tau_{max}-1$}{
    Gradient step: $\dot{\bbS} = \bchkS^{(\tau)} - \eta\nabla_\bbS \ccalL \left( f_{\bbTheta^{(t+1)}}(\bbX | \bchkS^{(\tau)}), \ccalY_{train} \right)$ \\
    Proximal on $\gamma(\bbS)$: $\ddot{\bbS} =  \mathrm{prox}_{\eta\lambda\|\cdot\|_1}(\dot{\bbS})$ \\
    Proximal on $d(\bbS, \barbS)$: $\dddot{\bbS} =  \mathrm{prox}_{\eta\alpha d(\cdot,\barbS)}(\ddot{\bbS})$ \\
    Projection step: $\bchkS^{(\tau + 1)} = \Pi_\ccalS(\dddot{\bbS})$
    }
    Set $\bbS^{(t+1)} = \bchkS^{(\tau_{max})}$
}
$\hbTheta = \bbTheta^{(T_{max})},\; \hbS = \bbS^{(T_{max})}$.
\caption{Robust GNN with graph denoising.}
\label{alg:robust_gnn}
\end{algorithm}

\section{Numerical Results}
In this section, we evaluate the performance of the proposed architecture in different real-world datasets and compare it with several state-of-the-art alternatives.
The code related to the implementation of our robust GNN, as well as the code for all the simulations presented in this paper, is available in GitHub\footnote{\url{https://github.com/vmtenorio/robust_gnn}}.

\vspace{2mm}
\noindent
\textbf{Experiment setup}.
We approach the task of node classification for three standard datasets: Cornell, Wisconsin, and Texas~\cite{Pei2020GeomGCN}. 
The three datasets capture relationships between different web pages with each node representing a web page, and edges denoting hyperlinks between them.
Regarding the perturbations, we create and destroy edges uniformly at random from the original graph.
To ensure that the sparsity of the graph remains constant, the creation and destruction of edges is performed by rewiring existing links. 
Then, the results reported in the figures are the mean accuracy over the test set of nodes for 50 independent realizations of the experiments, where each realization considers a different perturbed GSO $\barbS$ and a different GNN parameter initialization.

\vspace{2mm}
\noindent
\textbf{Baselines}.
The proposed robust architecture, labeled as ``RGCNH'' in the figures, is compared with several alternatives.
First, we consider its non-robust counterpart, which is given by implementing the recursion in~\eqref{eq:GNN} using $\barbS$ as the real GSO.
This baseline is labeled as ``GCNH'' in the legend of the figures.
We also consider two popular non-robust GNNs shown to perform well in a wide range of problems, namely the ``GCN''~\cite{kipf2016semi} and the ``GAT''~\cite{velickovic2018graph}.
Finally, as a robust state-of-the-art method, we consider the ``ProGNN'' architecture proposed in~\cite{jin2020graph}.

\begin{figure*}[!t]
    \begin{subfigure}[t]{0.32\textwidth}
        \centering
	  \begin{tikzpicture}[baseline,scale=.9]

\begin{axis}[
    title={Cornell Dataset},
    table/col sep=semicolon,
    xlabel={(a) Percentage of perturbed links},
    width=7cm,
    height=6cm,
    xmin={-0.01},
    xmax={0.16},
    xtick={0, .02, .05, .1, .15},
    xticklabels={0, .02, .05, .1, .15},
    ylabel={Accuracy over the test set of nodes},
    ymin={0.35},
    ymax={0.80},
    grid=major,
    legend style={
        nodes={scale=1.2, transform shape},
        at={(0,0.65)},
        anchor=north west},
    label style={font=\small}
    ]
    
    \pgfplotstableread{data/20231124-perts.csv}\pertError
    
    \addplot[blue, mark=*] table [x=Pert, y=Cornell-GCNH] {\pertError};
    \addplot[red, mark=*] table [x=Pert, y=Cornell-RGCNH-M1] {\pertError};
    \addplot[orange, mark=*] table [x=Pert, y=Cornell-GCNN] {\pertError};
    \addplot[green, mark=*] table [x=Pert, y=Cornell-ProGNN] {\pertError};
    \addplot[cyan, mark=*] table [x=Pert, y=Cornell-GAT] {\pertError};
    
    \legend{GCNH, RGCNH, GCN, ProGNN, GAT}
    
\end{axis}
\end{tikzpicture}
    \end{subfigure}
    \begin{subfigure}[t]{0.32\textwidth}
        \centering
	   \begin{tikzpicture}[baseline,scale=.9]

\begin{axis}[
    title={Wisconsin Dataset},
    table/col sep=semicolon,
    xlabel={(b) Percentage of perturbed links},
    width=7cm,
    height=6cm,
    xmin={-0.01},
    xmax={0.16},
    xtick={0, .02, .05, .1, .15},
    xticklabels={0, .02, .05, .1, .15},
    ylabel={Accuracy over the test set of nodes},
    ymin={0.45},
    ymax={0.85},
    grid=major,
    legend style={
        nodes={scale=1.2, transform shape},
        at={(0,0.7)},
        anchor=north west},
    label style={font=\small}
    ]
    
    \pgfplotstableread{data/20231124-perts.csv}\pertError
    
    \addplot[blue, mark=*] table [x=Pert, y=Wisconsin-GCNH] {\pertError};
    \addplot[red, mark=*] table [x=Pert, y=Wisconsin-RGCNH-M1] {\pertError};
    \addplot[orange, mark=*] table [x=Pert, y=Wisconsin-GCNN] {\pertError};
    \addplot[green, mark=*] table [x=Pert, y=Wisconsin-ProGNN] {\pertError};
    \addplot[cyan, mark=*] table [x=Pert, y=Wisconsin-GAT] {\pertError};
    
    \legend{GCNH, RGCNH, GCN, ProGNN, GAT}
    
\end{axis}
\end{tikzpicture}
    \end{subfigure}
    \begin{subfigure}[t]{0.32\textwidth}
        \centering
	   \begin{tikzpicture}[baseline,scale=.9]

\begin{axis}[
    title={Texas Dataset},
    table/col sep=semicolon,
    xlabel={(c) Percentage of nodes with perturbations},
    width=7cm,
    height=6cm,
    xmin={-0.01},
    xmax={0.71},
    xtick={0, .1, .3, .5, .7},
    xticklabels={0, .1, .3, .5, .7},
    ylabel={Accuracy over the test set of nodes},
    ymin={0.5},
    ymax={0.83},
    grid=major,
    legend style={
        nodes={scale=1.2, transform shape},
        at={(0,0.7)},
        anchor=north west},
    label style={font=\small}
    ]
    
    \pgfplotstableread{data/20231124-pertsSubset.csv}\pertError
    
    \addplot[blue, mark=*] table [x=Pert, y=Texas-GCNH] {\pertError};
    \addplot[red, mark=*] table [x=Pert, y=Texas-RGCNH-M1-Prior-T] {\pertError};
    \addplot[orange, mark=*] table [x=Pert, y=Texas-GCNN] {\pertError};
    \addplot[green, mark=*] table [x=Pert, y=Texas-ProGNN] {\pertError};
    \addplot[cyan, mark=*] table [x=Pert, y=Texas-GAT] {\pertError};
    
    \legend{GCNH, RGCNH, GCN, ProGNN, GAT}
    
\end{axis}
\end{tikzpicture}
    \end{subfigure}
	\caption{Mean accuracy of the node classification task for Cornell, Wisconsin, and Texas datasets in the presence of different types of perturbations. Perturbations in panels (a) and (b) consist in rewiring an increasing percentage of all edges of the graph, while in panel (c) perturbations only affect to edges between a subset of nodes.}\label{fig:pert_err}
\end{figure*}
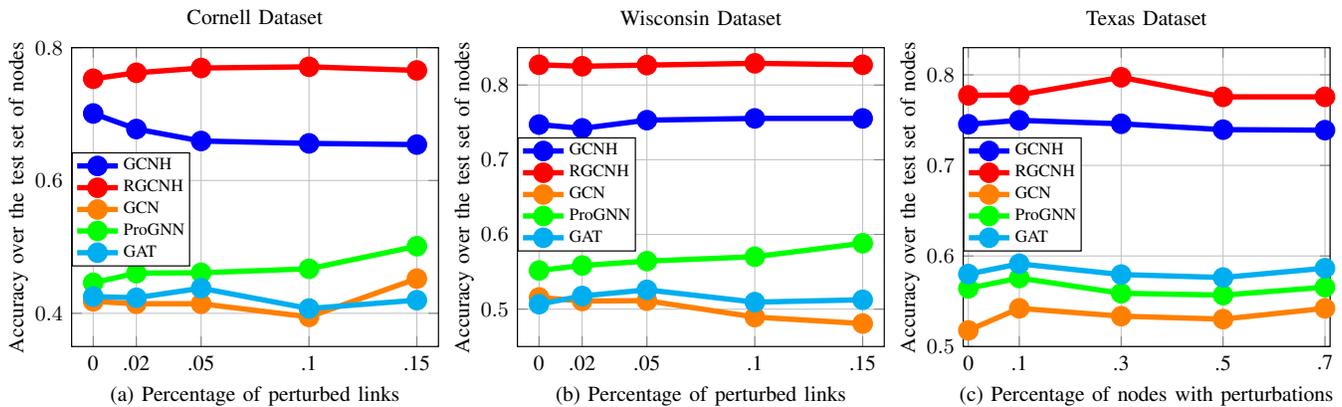

\vspace{2mm}
\noindent
\textbf{Test case 1}. The first experiment analyzes the performance of the proposed methodology as we increase the number of edges that are perturbed. Figure~\ref{fig:pert_err}~(a) and Figure~\ref{fig:pert_err}~(b) represent the mean accuracy over the test set of nodes for the Cornell and Wisconsin datasets, respectively, as we increase the perturbation probability as indicated in the x-axis.
From the results, it is clear that our proposed methodology consistently outperforms all the alternatives.
In Figure~\ref{fig:pert_err} (a) we observe that the accuracy of the ``RGCNH'' remains constant despite the increasing percentage of perturbed edges while the accuracy of the non-robust alternative ``GCNH'' deteriorates, highlighting the benefit of accounting for the influence of perturbations.
As for ``ProGNN'', it exhibits better performance than its non-robust counterpart, ``GCN''.
However, its low accuracy may be due to two factors.
First, the results suggest that the GCN is not an appropriate architecture for these datasets, and second, the assumptions about the true $\bbS$ do not need to hold in these datasets.

\vspace{2mm}
\noindent
\textbf{Test case 2}. 
In this case, we consider a perturbation with more structure and aim to analyze the benefits of including additional information about the perturbation.
To that end, we consider a perturbation that only affects edges between a subset of the nodes of the graph, and incorporate this prior information into the definition of $d (\bbS, \barbS)$ by only applying the proximal operator for the $\ell_1$ norm on the submatrix corresponding to the nodes with possibly perturbed edges.
Figure~\ref{fig:pert_err} (c) shows the accuracy for the test nodes in the Texas dataset as we increase the percentage of nodes with perturbations. We can see that, while the performance of the architecture that ignores perturbations decreases slightly, our architecture outperforms the alternatives and its accuracy remains constant as we increase the intensity of the perturbations.

\section{Conclusion}
This work introduced a novel GNN architecture robust to perturbations in the topology of the graph.
Assuming that only $\barbS$, a perturbed version of the GSO, is available, the key features of the proposed method are: i) considering $\bbS$, the true GSO, as an optimization variable, and ii) learn (optimize) jointly matrix $\bbS$ and $\bbTheta$, the weights of the GNN.
Since this is a challenging ill-posed problem, the optimization objective was augmented with prior knowledge about: i) the true GSO, and ii) the strength of the perturbations. 
Training the proposed GNN entails solving a challenging optimization problem, so we designed an iterative alternating minimization algorithm that, at each iteration, estimates sequentially $\bbTheta$ and $\bbS$. Moreover, to deal with the non-differentiable terms when estimating $\bbS$, we employed a projected proximal gradient descent algorithm.
Finally, we assessed the performance of our robust GNN in several datasets and compared it with state-of-the-art alternatives.


\bibliographystyle{IEEEbib}
\bibliography{myIEEEabrv,biblio}

\begin{thebibliography}{10}

\bibitem{scarselli2009gnn}
F.~Scarselli, M.~Gori, A.C. Tsoi, M.~Hagenbuchner, and G.~Monfardini,
\newblock ``The graph neural network model,''
\newblock {\em {IEEE} Trans. Neural Netw.}, vol. 20, no. 1, pp. 61--80, Jan.
  2009.

\bibitem{bronstein2017geometricdeeplearning}
M.M. {Bronstein}, J.~{Bruna}, Y.~{LeCun}, A.~{Szlam}, and P.~{Vandergheynst},
\newblock ``Geometric deep learning: Going beyond euclidean data,''
\newblock {\em IEEE Signal Process. Mag.}, vol. 34, no. 4, pp. 18--42, July
  2017.

\bibitem{wu2020comprehensive}
Z.~Wu, S.~Pan, F.~Chen, G.~Long, C.~Zhang, and S.~Y. Philip,
\newblock ``A comprehensive survey on graph neural networks,''
\newblock {\em IEEE Trans. Neural Netw. Learn. Syst.}, 2020.

\bibitem{bronstein2021geometric}
M.~M. Bronstein, J.~Bruna, T.~Cohen, and P.~Veli{\v{c}}kovi{\'c},
\newblock ``Geometric deep learning: Grids, groups, graphs, geodesics, and
  gauges,''
\newblock {\em arXiv preprint arXiv:2104.13478}, 2021.

\bibitem{rey2021overparametrized}
S.~Rey, V.~M. Tenorio, S.~Rozada, L.~Martino, and A.~G.~Marques,
\newblock ``Overparametrized deep encoder-decoder schemes for inputs and
  outputs defined over graphs,''
\newblock in {\em European Signal Process. Conf. (EUSIPCO)}. IEEE, 2021, pp.
  855--859.

\bibitem{kolaczyk2009book}
E.~D. Kolaczyk,
\newblock {\em Statistical Analysis of Network Data: Methods and Models},
\newblock Springer, New York, NY, 2009.

\bibitem{shuman2013emerging}
D.I. Shuman, S.K. Narang, P.~Frossard, A.~Ortega, and P.~Vandergheynst,
\newblock ``The emerging field of signal processing on graphs: Extending
  high-dimensional data analysis to networks and other irregular domains,''
\newblock {\em IEEE Signal Process. Mag.}, vol. 30, no. 3, pp. 83--98, 2013.

\bibitem{djuric2018cooperative}
P.~Djuric and C.~Richard,
\newblock {\em Cooperative and Graph Signal Processing: {P}rinciples and
  Applications},
\newblock Academic Press, 2018.

\bibitem{marques2020editorial}
A.~G. Marques, N.~Kiyavash, J.~M.~F. Moura, D.~Van~De Ville, and R.~Willett,
\newblock ``Graph signal processing: {F}oundations and emerging directions
  (editorial),''
\newblock {\em IEEE Signal Process. Mag.}, vol. 37, Nov. 2020.

\bibitem{kearnes2016molecular}
Steven Kearnes, Kevin McCloskey, Marc Berndl, Vijay Pande, and Patrick Riley,
\newblock ``Molecular graph convolutions: moving beyond fingerprints,''
\newblock {\em J. Computer-Aided Molecular Des.}, vol. 30, pp. 595--608, 2016.

\bibitem{ying2018graph}
R.~Ying, R.~He, K.~Chen, P.~Eksombatchai, W.~L. Hamilton, and J.~Leskovec,
\newblock ``Graph convolutional neural networks for web-scale recommender
  systems,''
\newblock in {\em Intl. Conf. Knowl. Discovery Data Mining (ACM SIGKDD)}, 2018,
  pp. 974--983.

\bibitem{guo2019attention}
S.~Guo, Y.~Lin, N.~Feng, C.~Song, and H.~Wan,
\newblock ``Attention based spatial-temporal graph convolutional networks for
  traffic flow forecasting,''
\newblock in {\em AAAI Conf. Artificial Intell.}, 2019, vol.~33, pp. 922--929.

\bibitem{zhou2020graph}
J.~Zhou, G.~Cui, S.~Hu, Z.~Zhang, C.~Yang, Z.~Liu, L.~Wang, C.~Li, and M.~Sun,
\newblock ``Graph neural networks: A review of methods and applications,''
\newblock {\em AI open}, vol. 1, pp. 57--81, 2020.

\bibitem{rey2022untrained}
S.~Rey, S.~Segarra, R.~Heckel, and A.~G. Marques,
\newblock ``Untrained graph neural networks for denoising,''
\newblock {\em IEEE Trans. Signal Process.}, vol. 70, pp. 5708--5723, 2022.

\bibitem{dai2018adversarial}
H.~Dai, H.~Li, T.~Tian, X.~Huang, L.~Wang, J.~Zhu, and L.~Song,
\newblock ``Adversarial attack on graph structured data,''
\newblock in {\em Intl. Conf. Machine Learn. (ICML)}, 10--15 Jul 2018, vol.~80,
  pp. 1115--1124.

\bibitem{ceci2020graph}
E.~Ceci and S.~Barbarossa,
\newblock ``Graph signal processing in the presence of topology
  uncertainties,''
\newblock {\em IEEE Trans. Signal Process.}, vol. 68, pp. 1558--1573, 2020.

\bibitem{kenlay2021stability}
H.~Kenlay, D.~Thano, and X.~Dong,
\newblock ``On the stability of graph convolutional neural networks under edge
  rewiring,''
\newblock in {\em IEEE Intl. Conf. Acoustics, Speech and Signal Process.},
  2021, pp. 8513--8517.

\bibitem{nguyen2022stability}
H.~S. Nguyen, Y.~He, and H.~T. Wai,
\newblock ``On the stability of low pass graph filter with a large number of
  edge rewires,''
\newblock in {\em IEEE Intl. Conf. Acoustics, Speech and Signal Process.},
  2022, pp. 5568--5572.

\bibitem{levie2021transferability}
R.~{Levie et al.},
\newblock ``Transferability of spectral graph convolutional neural networks.,''
\newblock {\em J. Mach. Learn. Res.}, vol. 22, pp. 272--1, 2021.

\bibitem{ruiz2021graph}
L.~Ruiz, F.~Gama, and A.~Ribeiro,
\newblock ``Graph neural networks: {A}rchitectures, stability, and
  transferability,''
\newblock {\em Proc. IEEE}, vol. 109, no. 5, pp. 660--682, 2021.

\bibitem{keriven2020convergence}
N.~Keriven, A.~Bietti, and S.~Vaiter,
\newblock ``Convergence and stability of graph convolutional networks on large
  random graphs,''
\newblock {\em Conf. Neural Inform. Process. Syst.}, vol. 33, pp. 21512--21523,
  2020.

\bibitem{rey2023robust}
S.~Rey, V.~M. Tenorio, and A.~G. Marques,
\newblock ``Robust graph filter identification and graph denoising from signal
  observations,''
\newblock {\em IEEE Trans. Signal Process.}, vol. 71, pp. 3651--3666, 2023.

\bibitem{ceci2020_semtls}
E.~Ceci, Y.~Shen, G.~B. Giannakis, and S.~Barbarossa,
\newblock ``Graph-based learning under perturbations via total least-squares,''
\newblock {\em IEEE Trans. Signal Process.}, vol. 68, pp. 2870--2882, 2020.

\bibitem{natali2020topology}
A.~Natali, M.~Coutino, and G.~Leus,
\newblock ``Topology-aware joint graph filter and edge weight identification
  for network processes,''
\newblock in {\em IEEE Intl. Wrkshp. Mach. Learn. Signal Process. (MLSP)}.
  IEEE, 2020, pp. 1--6.

\bibitem{rey2021robust}
S.~Rey and A.~G. Marques,
\newblock ``Robust graph-filter identification with graph denoising
  regularization,''
\newblock in {\em IEEE Intl. Conf. Acoustics, Speech and Signal Process.} IEEE,
  2021, pp. 5300--5304.

\bibitem{tenorio2021robust}
V.~M. Tenorio, S.~Rey, F.~Gama, S.~Segarra, and A.~G. Marques,
\newblock ``A robust alternative for graph convolutional neural networks via
  graph neighborhood filters,''
\newblock in {\em Asilomar Conf. Signals, Systems and Comput.} IEEE, 2021, pp.
  1573--1578.

\bibitem{jin2020graph}
W.~Jin, Y.~Ma, X.~Liu, X.~Tang, S.~Wang, and J.~Tang,
\newblock ``Graph structure learning for robust graph neural networks,''
\newblock in {\em Intl. Conf. Knowl. Discovery Data Mining (ACM SIGKDD)}, 2020,
  pp. 66--74.

\bibitem{kipf2016semi}
T.~N. Kipf and M.~Welling,
\newblock ``Semi-supervised classification with graph convolutional networks,''
\newblock {\em arXiv preprint arXiv:1609.02907}, 2016.

\bibitem{chen2020measuring}
D.~Chen, Y.~Lin, W.~Li, P.~Li, J.~Zhou, and X.~Sun,
\newblock ``Measuring and {Relieving} the {Over}-{Smoothing} {Problem} for
  {Graph} {Neural} {Networks} from the {Topological} {View},''
\newblock {\em AAAI Conf. Artificial Intell.}, vol. 34, no. 04, pp. 3438--3445,
  Apr. 2020,
\newblock Number: 04.

\bibitem{Pei2020GeomGCN}
H.~Pei, B.~Wei, K.~C. Chang, Y.~Lei, and B.~Yang,
\newblock ``Geom-gcn: Geometric graph convolutional networks,''
\newblock in {\em Intl. Conf. Learn. Representations (ICLR)}, 2020.

\bibitem{velickovic2018graph}
P.~Veli{\v{c}}kovi{\'{c}}, G.~Cucurull, A.~Casanova, A.~Romero, P.~Li{\`{o}},
  and Y.~Bengio,
\newblock ``{Graph Attention Networks},''
\newblock {\em Intl. Conf. Learn. Representations (ICLR)}, 2018.

\end{thebibliography}

\end{document}